# Super-resolution Probabilistic Rain Prediction from Satellite Data Using 3D U-Nets and EarthFormers

Yang Li[1,2], Haiyu Dong[*,2], Zuliang Fang[2], Jonathan Weyn[2], Pete Luferenko[2]

1. Nanjing University of Information Science and Technology
2. Microsoft

* Haiyu.Dong@microsoft.com

**Abstract** Accurate and timely rain prediction is crucial for decision making and is also a challenging task. This paper presents a solution which won the 2$^{nd}$ prize in the Weather4cast 2022 NeurIPS competition using 3D U-Nets and EarthFormers for 8-hour probabilistic rain prediction based on multi-band satellite images. The spatial context effect of the input satellite image has been deeply explored and optimal context range has been found. Based on the imbalanced rain distribution, we trained multiple models with different loss functions. To further improve the model performance, multi-model ensemble and threshold optimization were used to produce the final probabilistic rain prediction. Experiment results and leaderboard scores demonstrate that optimal spatial context, combined loss function, multi-model ensemble, and threshold optimization all provide modest model gain. A permutation test was used to analyze the effect of each satellite band on rain prediction, and results show that satellite bands signifying cloud-top phase (8.7 µm) and cloud-top height (10.8 and 13.4 µm) are the best predictors for rain prediction. The source code is available at https://github.com/bugsuse/weather4cast-2022-stage2.

**Keywords**: rain prediction, satellite data, super-resolution, 3D U-Net, EarthFormer, nowcasting

## 1. Introduction

Precipitation significantly impacts peoples' daily lives. Accurate and timely short-term precipitation prediction (0-6 hour) is crucial for weather-dependent decision-making [1, 2]. Broadly, precipitation prediction techniques can be divided into two classes: statistical modeling using radar echo and/or satellite image extrapolation, and numerical weather prediction (NWP) [3, 4]. NWP typically requires vast computational resources to integrate mathematical physical forecasting equations at high temporal and spatial resolutions, making it challenging for the desired frequent update cycles. The statistical extrapolation method can use near-real-time observations to forecast convection and precipitation with much lower latency. Therefore, radar and/or satellite image extrapolation convection nowcasting approaches have been used operationally in many national and regional weather centers [5].

Recent advances in machine learning and deep learning have greatly improved data inference ability in the diverse domains of business and science [6]. Machine learning and deep learning methods have been successfully employed to predict the spatiotemporal evolution of weather based on radar-image and/or satellite-image extrapolation [1,2,5,7,8,9], such as Weather4cast 2021 weather forecasting challenge [10,11]. To further drive innovation in the application of machine learning, the Weather4cast 2022 NeurIPS Competition[1] was held, with the aim of super-resolution rain movie prediction under spatio-temporal shifts. The competition task is to predict, in each region, the next 8 hours (32 time slots) of high-resolution rainfall rate from the OPERA radar network given the preceding hour (4 time slots) of coarser spatial resolution multi-band satellite measurements.

The Weather4cast 2022 competition has three stages: Stage1, Stage2 Core and Transfer Learning challenge. This paper mainly presents the solution that we are using in Stage2 Core and Transfer Learning challenge. We used 3D U-Net [12, 13] and EarthFormer [9] for this task. Our solution was to effectively leverage synoptic-scale and mesoscale context information of visible, near-infrared, water vapor and infrared bands variables. The data and methods are described in Section 2. The results are presented in section 3. Section 4 provides a brief discussion and conclusion.

## 2. Data and Methods

### 2.1 Satellite and rainfall rate data

The Weather4cast 2022 NeurIPS Competition presents a challenging problem of rain prediction: predict rainfall rates from time evolution of multi-band (i.e., visible, near-infrared (Near-IR), water vapor and infrared (IR) bands, in Table 1) satellite-based measurements. Multi-band satellite measurements are from the European Organization for the Exploitation of Meteorological Satellites (EUMETSAT) MeteoSat satellites. The rainfall rates are calculated based on ground-radar-reflectivity measurements by the Operational Program for Exchange of Weather Radar Information (OPERA) radar

---
[1] https://www.iarai.ac.at/weather4cast/

network.

Table 1. List of input satellite-based predictors were used for rain prediction, along with their band type, central wavelength and physical basis.

| Band type | Central wavelength (μm) | Physical basis |
|---|---|---|
| Visible | 0.64 | Vegetation, low clouds and fog |
| Near-IR | 0.81 | Vegetation and aerosol |
| | 1.6 | Snow/ice |
| IR | 3.9 | Low clouds and fog, wildfires |
| | 6.3 | Upper-level tropospheric water vapor |
| | 7.4 | Lower-level tropospheric water vapor |
| | 8.7 | Cloud-top phase |
| | 9.7 | Ozone |
| | 10.8 | "Clean" IR longwave window |
| | 12 | "Dirty" longwave window |
| | 13.4 | $CO_2$ longwave |

As shown in Figure 1[2], the multi-band satellite image input products and the OPERA ground-radar rainfall rates as target are given for 252×252 pixel patch but note that the spatial resolution of the satellite images (12 km) is about six times lower than the resolution of the ground-radar rain rate products (2 km). The ground-radar rainfall products with 252×252 pixel patch correspond to a 42×42 pixel patch centered region in the coarser satellite resolution. A large input area surrounding the target region can supply sufficient synoptic-scale and mesoscale context information for a prediction of future weather, particularly for long lead times [8].

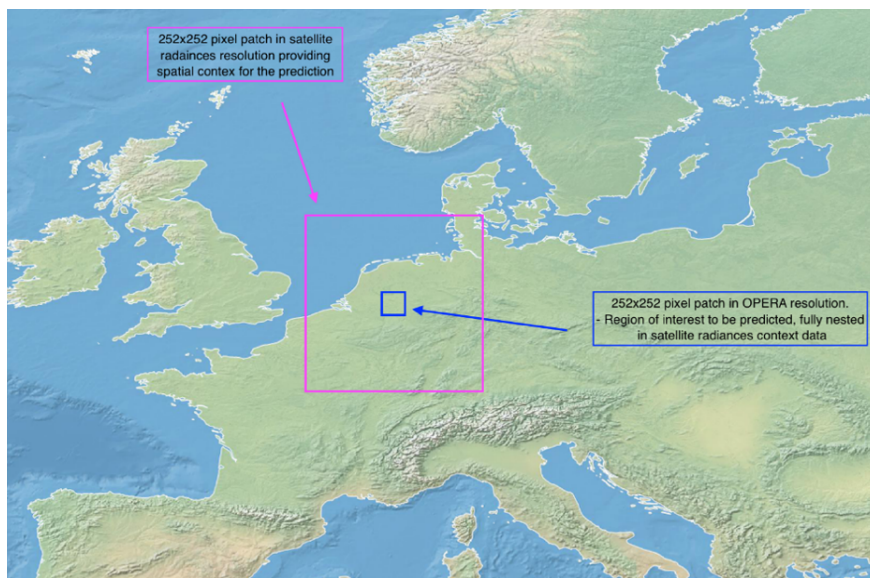

---

[2] https://github.com/iarai/weather4cast-2022

**Figure 1**. Illustration of input satellite-based images and target region context. The pink rectangle covers the input satellite radiances used to capture increasing amounts of context around the target patch (blue rectangle).

For Stage 1, only rain/no rain needs to be predicted for each pixel using satellite-based images and rainfall rate products from one year only, covering Feb. to Dec. 2019, on three European regions (boxi_0015, boxi_0034, boxi_0076). For the Stage 2 Core challenge, forecasting rain events of rain rate threshold 0.2 mm uses much more data, including all the data from Stage 1, and extending it to cover 7 regions (boxi_0015, boxi_0034, boxi_0076, roxi_0004, roxi_0005, roxi_0006, roxi_0007) with training data from 2 years (2019 and 2020). A total number of 228928 samples in training dataset and 840 samples in validation dataset were generated using a sliding window method. In the Stage 2 Transfer Learning challenge, test data of 3 additional regions (e.g., roxi_0008, roxi_0009, roxi_0010) in 3 years (2019-2021), and existing 7 regions in the third year (2021) were used to assess the temporal and/or spatial transfer learning. In addition, static data with the elevation of the terrain, longitude and latitude are available for each region.

## 2.2 Description of models and training strategy

### 2.2.1 Model architectures

During Stage 1, we tested a couple of neural network models that are used on spatial-temporal tasks, mainly three types: U-Nets (U-Net [14], 3D U-Net, and U$^2$Net [15]); RNNs (ConvLSTM [7] and trajGRU [1]); and Transformers (Swin Transformer [16], EarthFormer). Based on the experiment results of Stage 1 (described in Results section), 3D U-Net and EarthFormer were used on Stage 2.

As shown in Figure 2, the 3D U-Net model consists of five encoder blocks, four decoder blocks and one output block. The encoder block performs a 2x downsampling, including 3d convolutional layers, max-pooling layers, BatchNorm, ReLU activation function, and Dropout3d layer. The decoder block consists of 3d convolutional layers, upsampling layers via transposed convolution, BatchNorm and ReLU activation function. The convolutional layers at different depths can extract spatial features at different resolutions, which is crucial for precipitation prediction, due to the multiscale nature of weather phenomena [17]. Each pooling layer downsample the feature maps detected by convolutional layer to a lower spatial resolution. The Dropout layer helps prevent overfitting by randomly setting input units to 0 with a probability of 0.4 at each step during training time. The BatchNorm layer is a method used to make training of deep neural networks faster and stable by standardizing the inputs to a layer for each mini-batch [18].

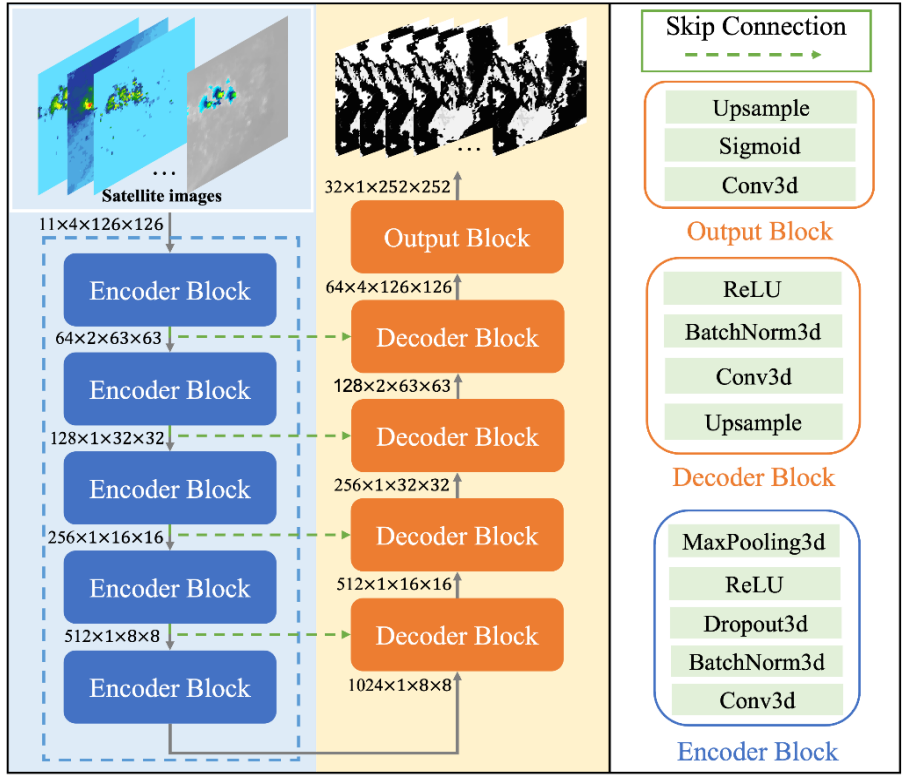

**Figure 2.** Illustration of the 3D U-Net model. The 3D U-Net model consists of five encoder blocks, four decoder blocks, one output block and skip connection.

Skip connections preserve high-resolution information by concatenating the feature maps from the encoder and decoder pathway along the channel axis, which is otherwise lost by downsampling. Feature maps concatenation generally leads to better model accuracy than directly add feature maps like in ResNet [19]. To generate precipitation probability prediction, the Output block contains a convolutional layer and an upsample layer with a scale of 2 and bilinear interpolation. A sigmoid activation function collapses the outputs to range from 0 to 1, yielding the final prediction of precipitation probability.

As shown in Figure 3a, the EarthFormer is a hierarchical transformer encoder-decoder based on cuboid attention. The cuboid attention layer involves three steps: "decompose", "attend" and "merge", shown in Figure 3b. First, the input spatiotemporal tensor is decomposed into a sequence of cuboids using local decomposition strategy or the dilated decomposition strategy. Second, the self-attention within each cuboid attention layer is applied in parallel to a sequence of non-overlapping cuboids from the decompose step. Finally, the sequence of cuboids obtained after the attention step are merged back to the original input shape to produce the final output of cuboid attention. The "merge" step is the inverse operation of "decompose" step. Due to the cuboids not communicating with each other, each cuboid is not capable of understanding the global dynamics of the system. Therefore, the elements will not only attend to the other elements within the same cuboid but also attend to the global vectors when each cuboid is performing the self-attention. More details about the EarthFormer can be found in [9]. To generate the precipitation probability prediction, we apply the sigmoid layer to make the outputs of the EarthFormer range from 0 to 1.

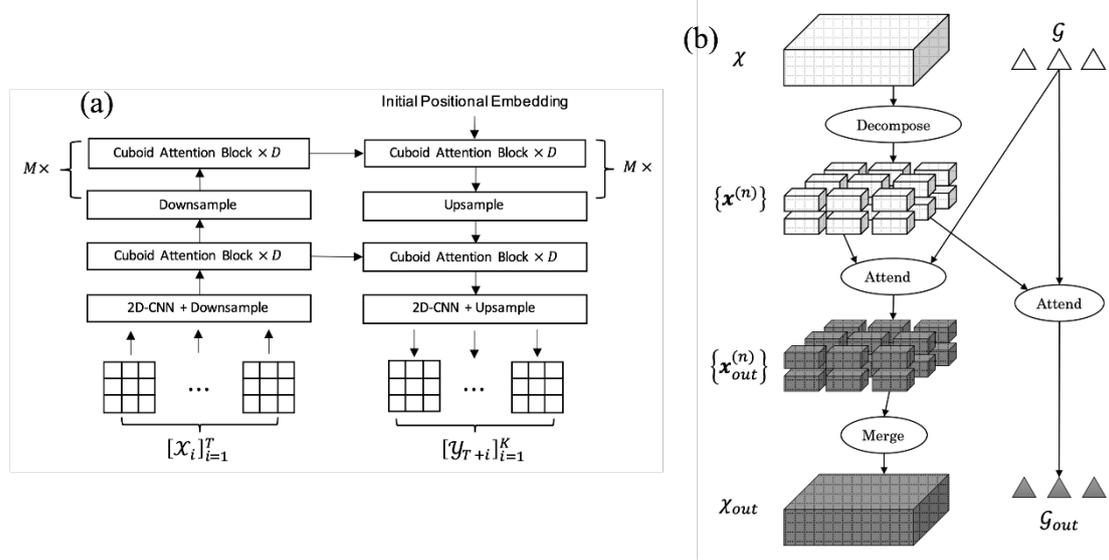

Figure 3. Illustration of (a) the EarthFormer architecture and (b) the cuboid attention layer with global vectors. The EarthFormer is a hierarchical Transformer encoder-decoder based on cuboid attention. The input sequence has length T and the target sequence has length K. "×D" means to stack D cuboid attention blocks with residual connection. "×M" means to have M layers of hierarchies. See Figure 2 and 3 of [9].

Considering the large dimensionality of this prediction task (e.g., N time frames of 11 bands of satellite images), we changed the start filter parameter from 32 to 64 and used large base units for the initial downsample stack layer in EarthFormer. The input multi-band satellite-based images were cropped to 126×126 and then fed into the 3D U-Net and EarthFormer to train for rainfall prediction, and an upsample layer with scale factor 2 was used to interpolate the output of the model to 252×252.

2.2.2 Training and optimization strategy

The 3D U-Net and EarthFormer models were trained on the training set of seven regions (i.e., boxi_0015, boxi_0034, boxi_0076, roxi_0004, roxi_0005, roxi_006 and roxi_0007) on the Core leaderboard using a sequence of 4 time frames of multi-band satellite-based measurements as input variables and a sequence of 32 time frames of rainfall rate by the OPERA radar network from 2019 to 2020 as target. The static data (elevation, longitude and latitude) were not used in model training for the final submission, as we found it did not improve model performance in experiments.

Large synoptic-scale context can carry useful circulation information for precipitation prediction in advance hours, particularly for heavy precipitation. Based on a characteristic synoptic-scale motion of 10 m/s [20, 21] and storm-motion velocity of ~15 m/s [22] over 8 hours, we select an input size encompassing a region 432 km larger than the target prediction region. Finally, 126×126 was used as the final input size in order to provide more synoptic-scale information. Therefore, we cropped the input satellite images from 252×252 to 126×126 as the final input variables fed into the 3D

U-Net and EarthFormer models. In addition, to ensure stable model training, the samples of all regions were shuffled before training to have samples of each batch are from multiple regions, not one region.

The 3D U-Net and EarthFormer models were trained with gradient clipping strategy using different loss functions such as IoU (Intersection over Union is calculated by dividing the overlap between the predicted and ground truth annotation by the union of these.), Dice, Focal and their combination (i.e., IoUDice and IoUDiceFocal). Table 2 shows the description of parameters used for training the 3D U-Net and EarthFormer models. The 3D U-Net and EarthFormer were both trained using the Adam [23] optimizer with the weight decay set to 5e-6, but with different learning rate and batch size. The learning rate was reduced by a factor of 2 whenever the model showed no improvement in validation loss for two consecutive epochs. In addition, if the validation loss did not improve in 10 epochs, the training would be stopped early. One NVIDIA V100 GPU with 32GB memory was used to train the 3D U-Net and EarthFormer models. Training for each epoch took approximately 90-120 min. More details about the model training parameters are available in the configuration files in our source code.

Table 2. Description of parameters used for training the 3D U-Net and EarthFormer models.

|  | 3D U-Net | EarthFormer |
|---|---|---|
| parameters (M) | 90.3 | 9.7/10.5 |
| optimizer | Adam | Adam |
| scheduler | ReduceLROnPlateau | ReduceLROnPlateau |
| learning rate | 8e-4 | 1e-4 |
| batch size | 64 | 12 |
| epochs | 90 | 90 |

The 3D U-Net model converges within about 30-50 epochs depending on the loss function and learning rate, while EarthFormer takes only 3-15 epochs. The best model checkpoints were selected based on validation-set metrics.

## 3. Results

**3.1 Sensitivity to the synoptic-scale and mesoscale context information**

As shown in Table 3, 126×126 pixels input satellite-based image has the best score on validation and leaderboard testing dataset. This result is consistent with our expectation of the large synoptic-scale and mesoscale context characteristic described in Section 2.2.2. This shows that cropping of input satellite image size can both speed up model training and further improve model performance.

Table 3. Evaluation metrics (IoU) of 3D U-Net model trained by Dice loss function using satellite-based images with different input size on validation dataset and on leaderboard of Stage1.

| | 252×252 | 126×126 | 42×42 |
|---|---|---|---|
| Validation dataset | 0.3358 | **0.4007** | 0.2607 |
| Leaderboard | 0.3842 | **0.4083** | 0.3484 |

Figure 4 revealed that the IoU score with input size of 42×42 is better than 126×126 and 252×252 at short lead time (about 90 min of 6 frames), but the performance of input size of 42×42 dramatically dropped at long lead times (after 3 hours). Though the large input size of 126×126 is a little bit worse than 42×42 within 90 min, it performs significantly better than 42×42 at long lead times. 126×126 is also much better than 252×252 for all lead times.

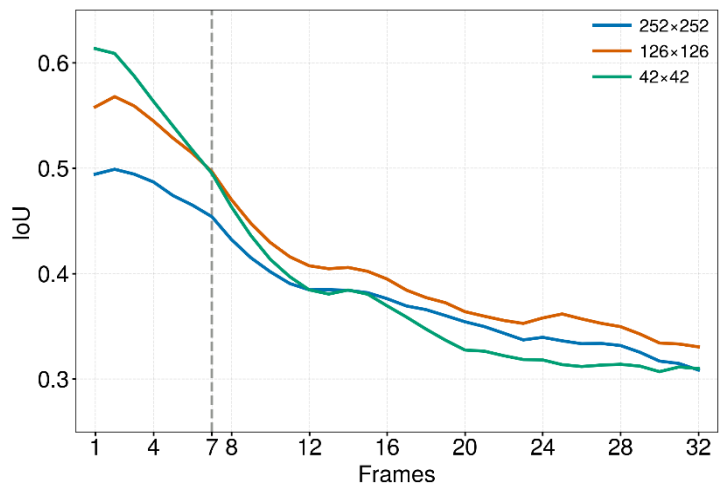

**Figure 4**. IoU time series on the validation dataset in Stage1 for 8-hour (32 frames) rain prediction using the 3D U-Net model with different input size.

This result demonstrates that mesoscale and local scale information in and around the target region is more favorable to short lead time (e.g., 2-hour) precipitation nowcasting than large synoptic-scale circulation context information. This indicates that using the large input size with more synoptic-scale context may not be beneficial for precipitation nowcasting with short lead time. This is especially true when the neural network model cannot effectively extract features that are favorable for prediction from large input context with rich synoptic-scale and mesoscale information. For long lead time precipitation prediction (e.g., 8-hour), large input image can supply more synoptic-scale and mesoscale background information. Therefore, for long lead time precipitation prediction, an optimal spatial context including enough large synoptic-scale and mesoscale context information should be further explored to make the best precipitation prediction.

### 3.2 The effects of loss function on rainfall rate prediction

In Stage2, two combined loss functions (IoUDice and IoUDiceFocal) were used to train the 3D U-Net and EarthFormer models. These losses were selected based on the Stage1 experiments of all loss functions (e.g., BCE, IoU, Dice, Focal and their combination). Table 4 shows the best IoU score of 3D U-Net and EarthFormer models that were trained with different loss functions (all have 126×126 input satellite-based images for the validation dataset on Stage2).

**Table 4.** Evaluation metrics (IoU) of the 3D U-Net and EarthFormer models using satellite-based images with 126×126 pixels on the validation dataset of Stage2.

|  | IoUDice | IoUDiceFocal |
| --- | --- | --- |
| 3D U-Net | 0.3003 | **0.3021** |
| EarthFormer | **0.3001** | 0.2895 |

### 3.3 Post-processing: ensemble and threshold optimization

With several candidate models of different architectures and loss functions showing good performance, a natural next step might be to try an ensemble prediction. We ensembled four models: two 3D U-Net models with IoUDice and IoUDiceFocal, and two different training cycles of the EarthFormer with IoUDice, shown in Table 5. The two different training cycles of the EarthFormer (EarthFormer_v1 and EarthFormer_v2) adopted the base units [16, 64, 128] and [64, 128, 128] for the initial downsample stack layers, respectively.

Direct averaging of probability prediction of the abovementioned four models is used as the final probability prediction of rainfall larger than 0.2 mm on Stage2. In the meanwhile, optimal threshold range was searched for each region to further improve the performance. Figure 5 shows the metric (IoU) distribution with probability thresholds at each region on the Core test leaderboard of Stage 2. The result revealed that the best probability threshold for different regions is different. IoU of ensemble average probability prediction with threshold optimization improved by 0.0114 over the best individual model on the Core testing leaderboard on Stage 2.

**Table 5.** Evaluation metrics (IoU) of the 3D U-Net and EarthFormer ensemble prediction on the Core testing leaderboard of Stage2.

|  | 3D U-Net IoUDice | 3D U-Net IoUDiceFocal | EarthFormer_v1 IoUDice | EarthFormer_v2 IoUDice | Ensemble |
| --- | --- | --- | --- | --- | --- |
| IoU | 0.2691 | 0.2768 | 0.2559 | 0.2667 | **0.2882** |

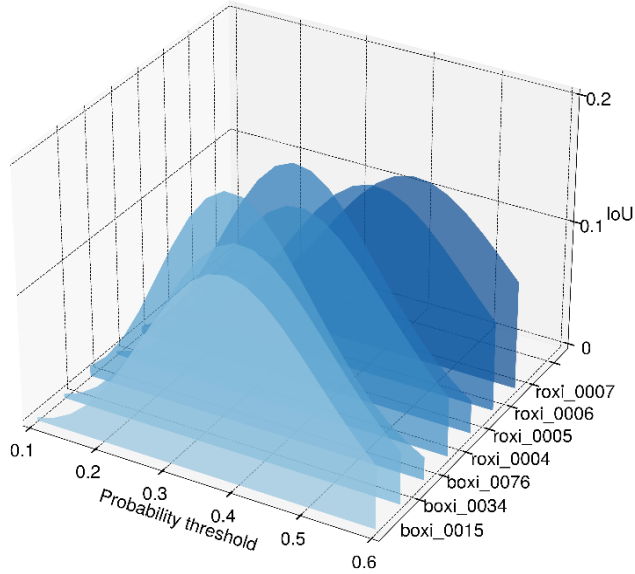

**Figure 5.** The metric distribution at different probability thresholds for each region on the Core test leaderboard of Stage 2.

Based on the evaluation results of the offline validation dataset and online Core test leaderboard of Stage 2, three submissions using multi-model ensemble with different probability thresholds for each region were produced for the final Core and Transfer Learning challenge heldout leaderboard. The final scores are shown in Table 6. The difference across three submissions were not obvious on the Core and Transfer Learning heldout leaderboard, indicating that the predictions from the multi-model ensemble with probability threshold optimization not only had better metric but were more stable.

**Table 6.** Evaluation metrics of the 3D U-Net and EarthFormer ensemble prediction on the core and transfer learning heldout leaderboard of Stage2.

|  | Core | Transfer Learning |
| --- | --- | --- |
| Ensemble | 0.3076 | 0.2659 |

The example prediction in Figure 6 suggests that the 3D U-Net and EarthFormer have good forecasting skills for continuous and widespread precipitation. However, it's difficult to accurately predict local small-scale precipitation (e.g., orange circles), especially for longer lead times, such as after 3 hours. Therefore, in addition to providing important synoptic-scale context via coarser-resolution input satellite images, high-resolution meso- and micro- scale data should be supplied to carry out much more useful information for prediction of small-scale local precipitation events.

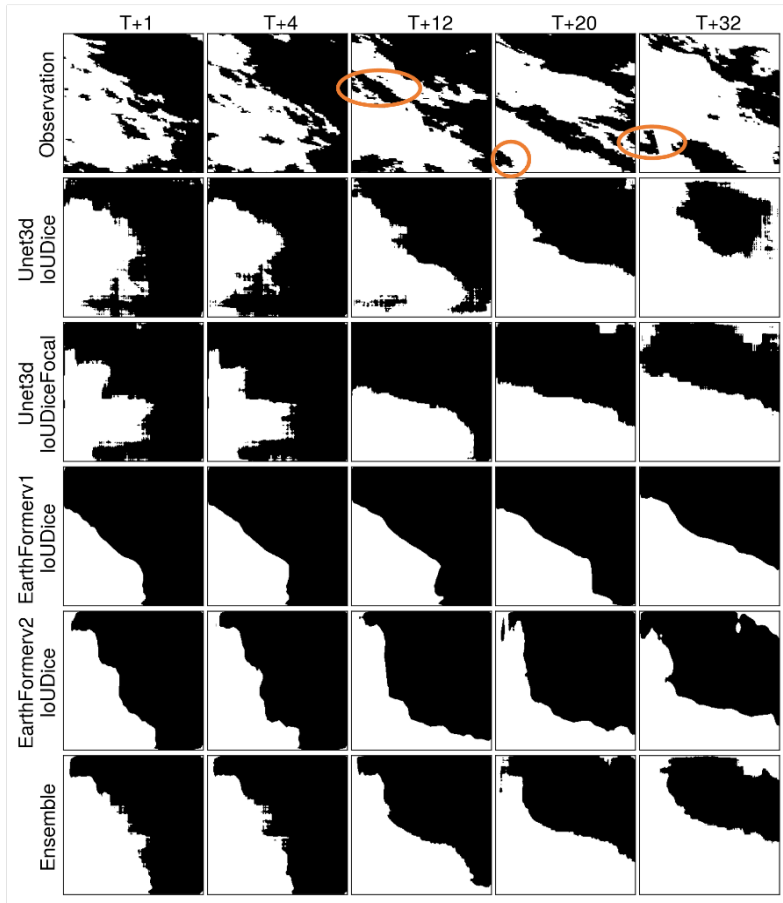

**Figure 6.** An example of rainfall observation and the 3D U-Net and EarthFormer prediction and ensemble average from the validation dataset in Stage2. The $T$ coordinate refers to the index of the frame in the sequence, where $T = 1$ is the first prediction. Orange circles indicate small-scale features of interest.

### 3.4 Effects of different input satellite bands

A permutation test [24] is a method for ranking the input predictors according to their importance to the metric (e.g., IoU). We randomized one input satellite band of all samples and all frames on the validation dataset at a time, then we evaluated on the validation dataset. The randomized input satellite band has a positive effect on precipitation prediction if the metric decreases. The more the metric drops, the more important the randomized input feature is for the prediction. Figure 7 shows the result of permutation test on each input satellite radiance band for rain prediction.

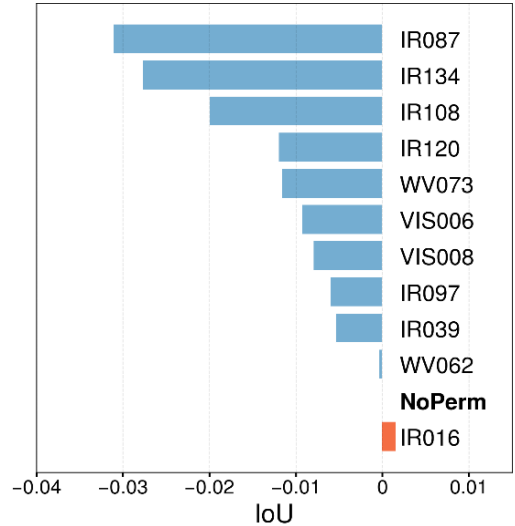

**Figure 7**. The result of permutation test on each input satellite radiance band for rain prediction.

The result demonstrates that input satellite band which signifies cloud-top phase (8.7 μm) and cloud-top height (10.8 and 13.4 μm) are the best for precipitation prediction. The dirty window band (12 μm), water vapor bands (especially lower-level tropospheric water vapor band of 7.3 μm and visible bands (VIS006 and VIS008) are also important for precipitation prediction on all regions. The satellite radiance band (6.2 μm) that indicates upper-level tropospheric water vapor has little influence on rain/no-rain binary prediction, but might be more useful for rainfall rate. Interestingly, the near-infrared band (IR016) that is used to distinguish snow/ice has a negative effect on rain prediction, but it has a positive effect on some regions, hence we continue to include it in our models.

## 4. Discussion and conclusion

The modeling solution presented above won the 2$^{nd}$ prize on both the Core and Transfer Learning leaderboards of the Weather4cast 2022 Challenge Stage2.

For short lead times (within 2 hours), input spatial context of 42×42 that equals the target context has the better performance, while for long lead hours (3~8 hours), input context of 126×126 that is 3 times of the target context performs better. From this, we can learn that mesoscale and local scale information is crucial for short lead times, while larger synoptic-scale and mesoscale context is a must for long lead times. Therefore, optimal input context for different lead times should be further explored with more experiments.

Imbalanced heavy rain distribution and different regional pattern are the key challenges in this competition task. Our combined loss function (i.e., IoUDice and IoUDiceFocal) showed great effectiveness on the imbalanced rain data problem. The multi-model ensemble with threshold optimization greatly improved the accuracy of rain prediction and, more importantly, the model generalization capability on different regions. This can be easily seen from the stable performance on offline evaluation, Core leaderboard, and Transfer Learning leaderboard.

In permutation test of each input satellite band on rain prediction, satellite bands of cloud-top phase (8.7 μm) and cloud-top height (10.8 and 13.4 μm) showed the highest importance for rain prediction, and satellite bands of upper-level tropospheric water vapor (6.2 μm) and snow/ice indicator (1.6 μm) have a little influence on the prediction.

One limitation we found is that all models cannot predict well on the local small-scale precipitation, especially for long lead times (after 3 hours). And we believe that local scales in convective events could be further improved by fusing multi-scale (large synoptic-scale and high resolution local small-scale) and multi-source (satellite, weather radar and dense automatic weather station) data.

# Reference

[1] Shi, X., Gao, Z., Lausen, L., Wang, H., Yeung, D. Y., Wong, W. K., and Woo, W. C. 2017. Deep learning for precipitation nowcasting: A benchmark and a new model. Advances in neural information processing systems, 30.

[2] Ravuri, S., and Coauthors, 2021. Skilful precipitation nowcasting using deep generative models of radar. Nature, 597(7878), 672-677.

[3] Dixon, M., and G. Wiener, 1993. TITAN: Thunderstorm identification, tracking, analysis, and nowcasting—A radar-based methodology. J. Atmos. Oceanic Technol., 10(6), 785−797.

[4] Sun, J. Z., and Coauthors, 2014. Use of NWP for nowcasting convective precipitation: recent progress and challenges. Bull. Amer. Meteor. Soc., 95(3), 409−426.

[5] Li, Y., Y. B. Liu, R. F. Sun, F. X. Guo, X. F. Xu, and H. X. Xu, 2023. Convective storm VIL and lightning nowcasting using satellite and weather radar measurements based on multi-task learning models. Adv. Atmos. Sci., https://doi.org/10.1007/s00376-022-2082-6

[6] LeCun, Y., Y., Bengio, and G. Hinton, 2015. Deep learning. Nature, 521(7553), 436−444, https://doi.org/10.1038/ nature14539.

[7] Shi, X. J., Z. R. Chen, H. Wang, D.-Y. Yeung, W.-K. Wong, and W.-C. Woo, 2015. Convolutional LSTM network: A machine learning approach for precipitation nowcasting. Proc. 28th International Conf. on Neural Information Processing Systems, Montreal, Canada, MIT Press, 802−810.

[8] Espeholt, L., Agrawal, S., Sønderby, et al., 2022. Deep learning for twelve hour precipitation forecasts. Nature communications, 13(1), 1-10.

[9] Gao, Z., Shi, X., Wang, H., Zhu, Y., Wang, Y., Li, M., & Yeung, D. Y., 2022. Earthformer: exploring space-time transformers for earth system forecasting. arXiv preprint arXiv:2207.05833.

[10] Gruca, A., Herruzo, P., Rípodas, P., Kucik, A., Briese, C., Kopp, M. K., ... & Kreil, D. P., 2021. CDCEO'21-First Workshop on Complex Data Challenges in Earth Observation. In Proceedings of the 30th ACM International Conference on Information & Knowledge Management (pp. 4878-4879).

[11] Herruzo, P., Gruca, A., Lliso, L., Calbet, X., Rípodas, P., Hochreiter, S., ... & Kreil, D. P., 2021. High-resolution multi-channel weather forecasting–First insights on transfer learning from the Weather4cast Competitions 2021. In 2021 IEEE International Conference on Big Data (Big Data) (pp. 5750-5757). IEEE.

[12] Çiçek, Ö., Abdulkadir, A., Lienkamp, S. S., Brox, T., & Ronneberger, O. (2016, October). 3D U-Net: learning dense volumetric segmentation from sparse annotation. In International conference on medical image computing and computer-assisted intervention (pp. 424-432). Springer, Cham.

[13] https://github.com/iarai/weather4cast-2022

[14] Ronneberger, O., Fischer, P., & Brox, T. (2015, October). U-net: Convolutional networks for biomedical image segmentation. In International Conference on Medical image computing and computer-assisted intervention (pp. 234-241). Springer, Cham.

[15] Qin, X., Zhang, Z., Huang, C., Dehghan, M., Zaiane, O. R., & Jagersand, M. (2020). U2-Net: Going deeper with nested U-structure for salient object detection. Pattern recognition, 106, 107404.

[16] Liu, Z., Lin, Y., Cao, Y., Hu, H., Wei, Y., Zhang, Z., ... & Guo, B. (2021). Swin transformer: Hierarchical vision transformer using shifted windows. In Proceedings of the IEEE/CVF International Conference on Computer Vision (pp. 10012-10022).

[17] Lagerquist, R., J. Q. Stewart, I. Ebert-Uphoff, and C. Kumler, 2021. Using Deep Learning to

Nowcast the Spatial Coverage of Convection from Himawari-8 Satellite Data. Mon. Wea. Rev., 149(12), 3897-3921.

[18] Ioffe, S., Szegedy, C., 2015. Batch normalization: Accelerating deep network training by reducing internal covariate shift. In International conference on machine learning (pp. 448-456). PMLR.

[19] Kayalibay, B., Jensen, G., van der Smagt, P., 2017. CNN-based segmentation of medical imaging data. arXiv preprint arXiv:1701.03056.

[20] Holton, J. R., & Hakim, G. J., 2012. An introduction to dynamic meteorology (Vol. 88). B.V.: Academic Press.

[21] Pan, B., Hsu, K., AghaKouchak, A., and Sorooshian, S., 2019. Improving precipitation estimation using convolutional neural network. Water Resources Research, 55(3), 2301-2321.

[22] Wapler, K., 2021. Mesocyclonic and non-mesocyclonic convective storms in Germany: Storm characteristics and life-cycle. Atmospheric Research, 248, 105186.

[23] D. P. Kingma, J. Ba, 2014. Adam: A method for stochastic optimization, in: 3rd International Conference for Learning Representations, San Diego, California, USA, URL: https://arxiv.org/abs/1412.6980.

[24] Breiman, L., 2001. Random forests. Machine learning, 45(1), 5-32.